\documentclass[sigconf]{acmart}
\usepackage{paralist}
\usepackage{float}
\restylefloat{table}
\usepackage{tabularx,booktabs}
\usepackage{multirow}%
\usepackage{balance}
\newcolumntype{C}{>{\centering\arraybackslash}X} 
\usepackage{lipsum} 
\usepackage{algorithm}
\usepackage{algorithmicx}
\usepackage{algpseudocode}
\usepackage{threeparttable} 
\usepackage{comment}
\usepackage{caption}
\usepackage{subcaption}
\usepackage{array}

\usepackage{varwidth}
\usepackage{enumitem}

\algnewcommand{\concatenate}{\text{Concatenate}}
\algnewcommand{\withElements}{\text{with elements of}}
\algnewcommand{\LineComment}[1]{\State \(\triangleright\) #1}
\AtBeginDocument{%
  \providecommand\BibTeX{{%
    \normalfont B\kern-0.5em{\scshape i\kern-0.25em b}\kern-0.8em\TeX}}}

\begin{document}
\fancyhead{}

\title{An Experiment in Retrofitting Competency Questions for Existing Ontologies }

\author{Reham Alharbi}
\orcid{0000-0002-8332-3803}
\affiliation{%
  \department{Department of Computer Science}
  \institution{University of Liverpool}
  \city{Liverpool}
  \country{UK}}
 \email{R.Alharbi@liverpool.ac.uk}

\author{Valentina Tamma}
\orcid{0000-0002-1320-610X}
\affiliation{%
  \department{Department of Computer Science}
  \institution{University of Liverpool}
  \city{Liverpool}
  \country{UK}}
 \email{V.Tamma@liverpool.ac.uk}
 
 \author{Floriana Grasso}
\orcid{0000-0001-8419-6554}
\affiliation{%
  \department{Department of Computer Science}
  \institution{University of Liverpool}
  \city{Liverpool}
  \country{UK}}
 \email{F.Grasso@liverpool.ac.uk}

 \author{Terry Payne}
\orcid{0000-0002-0106-8731}
\affiliation{%
  \department{Department of Computer Science}
  \institution{University of Liverpool}
  \city{Liverpool}
  \country{UK}}
 \email{T.R.Payne@liverpool.ac.uk}

\newcommand{\RA}[1]{\todo[inline, color=green!40]{RA:{\small #1}} }
\newcommand{\vt}[1]{\todo[inline, color=cyan!30]{VT:{ \small #1}} }
\newcommand{\nf}[1]{\todo[inline, color=orange!30]{FG:{ \small #1}} }
\renewcommand{\shortauthors}{R. Alharbi et al.}
\newtheorem{finding}{Finding}
\newtheorem{question}{Question}
\newcommand{\mysubsect}[1]{\vspace{0.0in}\noindent{\bf #1:}}

\begin{abstract}


Competency Questions (CQs) are a form of ontology functional requirements expressed as natural language questions. Inspecting CQs together with the axioms in an ontology provides critical insights into the intended scope and applicability of the ontology. CQs also underpin a number of tasks in the development of ontologies e.g. ontology reuse, ontology testing, requirement specification, and the definition of patterns that implement such requirements. Although CQs are integral to the majority of ontology engineering methodologies, the practice of publishing CQs alongside the ontological artefacts is not widely observed by the community. 

In this context, we present an experiment in retrofitting CQs from existing ontologies. We propose \textit{RETROFIT-CQs}, a method to extract candidate CQs directly from ontologies using Generative AI. In the paper we present the pipeline that facilitates the extraction of CQs by leveraging Large Language Models (LLMs) and we discuss its application to a number of existing ontologies.

\end{abstract}

\ccsdesc[500]{Computing methodologies~Natural language generation}

\keywords{Ontology Engineering, Competency Questions, Large Language Models, Reusability, Retrofitting}

\maketitle



\section{Introduction} \label{sec:introduction}

Competency Questions (\emph{CQs})~\cite{gruninger1995methodology} 
are a cornerstone of many ontology engineering methodologies~\cite{POVEDAVILLALON2022LOT,noy2001ontology}, as they
capture the tasks that arise in enterprise engineering and the consequent requirements that must be satisfied by the resulting ontology.
CQs are natural language questions and their related answers that an ontology-based application must be able to answer correctly, thereby ensuring that the resulting system has the relevant knowledge to successfully achieve its intended purpose. They not only serve as a litmus test for the development and evaluation of an ontology~\cite{noy2001ontology}, but also provide a common solution for modelling functional requirements~\cite{DBLP:conf/i-semantics/GangemiLLN22} in most traditional and agile ontology engineering methodologies; such as Pay as you go~\cite{juan2019}, NeOn~\cite{SurezFigueroa2015NEON}, eXtreme Design~\cite{Presutti2009eXtremeDW}, Ontology Development 101~\cite{noy2001ontology}, On-To-Knowledge~\cite{Staab2001} and LOT~\cite{POVEDAVILLALON2022LOT}.
Furthermore, they support the verification and evaluation~\cite{Keettestdriven2016,Bezerra2017} of the ontological artefact being built as their ``answerability'' becomes a functional requirement~\cite{Ren2014}.


Although several guidelines and methodologies (e.g. MIRO~\cite{adee4fe404784a1a955f5d15beb11e81} or LOT~\cite{POVEDAVILLALON2022LOT}) recommend that CQs are made available alongside an ontology artefact, they are often not published as part of the ontology documentation~\cite{coral, wisniewski2019analysis}, despite this being identified as a limitation both in the literature~\cite{Keettestdriven2016,Oscar2022,POVEDAVILLALON2022LOT} and by the practitioners themselves~\cite{alharbiKcap2021}. This can be problematic for third party developers who need to assess the correctness and reusability of an ontology.
In such cases, they often have to resort to manually inspecting the classes and properties of an ontology to decide whether it should be reused; this is a major limitation~\cite{alharbiKcap2021} as it can be highly subjective and is based on their level of expertise in ontology development.
Some approaches have emerged that assess ontologies for their reusability with respect to a set of requirements~\cite{Alharbi2021AssessingCO,Aziz2023}; however, they rely on the availability of CQs in addition to the ontology itself. 
Thus, a mechanism by which CQs could be elicited from a corresponding ontology would greatly facilitate its reusability.

In this paper, we explore the feasibility of retrofitting CQs from existing ontologies, by using the class and property labels in order to generate usable questions that should capture the intended scope of the ontology. To achieve this, we exploit recent advances of Large Language Models (LLMs) using tools such as ChatGPT\footnote{\url{https://openai.com/chatgpt}} and LLaMA\footnote{\url{https://github.com/facebookresearch/llama}}. Thus, our contribution addresses the following research question: \textbf{Q1: To what extent it is possible to generate ``usable'' CQs from existing ontologies that are representative of the scope and tasks for which the ontology was designed?} Here by ``usable'', we mean CQs that could be used in place of the original ones written by the ontology engineers (henceforth called \textit{design CQs})
for new tasks, e.g. assessing reusability wrt new requirements.
%
%
%
%
We present a pipeline that parses an ontology to extract relevant information and uses it to instantiate a prompt for the automatic generation of candidate CQs by an LLM. The validity of our proposal for candidate CQs generation is then evaluated by conducting two experiments. In the first,  the accuracy of the generated CQs is assessed with respect to the published CQs for the same ontology using CORAL~\cite{coral},
a comprehensive repository of CQs together 
with a dedicated CQ dataset \cite{POTONIEC2020105098}.
In the second, we solicit an ontology developer for an assessment of the CQs generated from one of their ontologies, and we ask them specifically to identify any questions that in hindsight would have been useful CQs at the time of designing the ontology.
The results confirm the hypothesis that generative models can be used to retrofit usable CQs. 



 
The paper is structured as follows: our research is framed with a discussion on the background in Section~\ref{sec:Background}, and the proposed approach is presented in Section~\ref{sec:approach}. 
Section~\ref{sec:Evaluation} presents the empirical evaluation whilst results are discussed in Section~\ref{sec:discussion}. Finally, the conclusions and future trends are outlined in Section~\ref{sec:conclusions}.


\section{Background} \label{sec:Background}

Several efforts have investigated approaches that identify CQ templates and patterns, or define a \emph{Controlled Natural Language} that facilitates their formalisation into a target logic or query language~\cite{Keettestdriven2016}.
Both ~\citeauthor{Ren2014}~\cite{Ren2014} and \citeauthor{Bezerra2017}~\cite{Bezerra2017} analyse a set of ontologies to identify CQ patterns. \citeauthor{Ren2014}~\cite{Ren2014} define a set of CQ ``archetypes'', syntactic patterns of CQs to be filled by domain expert; e.g.,  ``\texttt{Which [CE1] [OPE] [CE2]?}'', 
where \texttt{CE1} and \texttt{CE2} are class expressions (or individuals, in certain cases), and \texttt{OPE} is an object property expression. These patterns support ontology engineers in the formulation of machine processable CQs that can be used for ontology testing. Similarly, \citeauthor{Bezerra2017}~\cite{Bezerra2017} identify several patterns that can be instantiated with elements from the ontology vocabulary in order to specify CQs that can be used for the automatic testing and validation of an ontology.

\citeauthor{wisniewski2019analysis}~\cite{wisniewski2019analysis} catalogued 106 distinct CQ patterns determined over the CQs of five ontologies: Software Ontology (SWO), Stuff Ontology (Stuff), African Wildlife Ontology (AWO), Dementia Ambient Care ontology (Dem@Care), and  Ontology of Datatypes (OntoDT)~\cite{wisniewski2019analysis}. These were analysed and organised in different categories for each of the ontologies. \citeauthor{CLARO2019}~\cite{CLARO2019} built on 
this work with the proposal of CLaRO, a template-based controlled natural language to author CQs with 93 templates and their 41 variants.


Recent advances in Large Language Models (LLMs) have positioned them as a promising approach for the automatic generation of questions~\cite{DBLP:journals/pai/MullaG23} in natural language.
Auto-regressive LLMs (such as the GPT family~\cite{DBLP:conf/nips/Ouyang0JAWMZASR22}) are deep learning models trained on huge corpora of data in order to predict the next word in a sequence, given all of the previous words encountered. In particular, a novel paradigm for text generation is emerging, where LLMs are given a `prompt' in order to generate a desired output~\cite{DBLP:journals/pai/MullaG23}. In this context, a prompt consists of prepending a string to the context given to the LLMs~\cite{DBLP:journals/csur/LiuYFJHN23}, which includes some control element such as a keyword etc., to guide the generation of the text.

\section{The \emph{RETROFIT-CQ} pipeline} \label{sec:approach}

\begin{figure*}[t]
  \centering
  \includegraphics[width=0.8\linewidth]{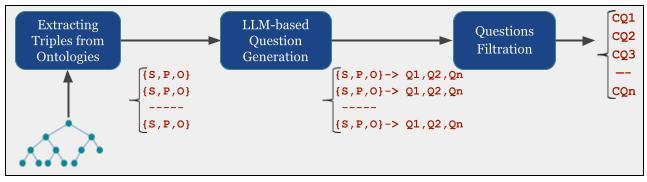}
  \vspace{-0.2in}
  \caption{The RETROFIT-CQs architecture.}
    \label{fig:approach}
\end{figure*}


The \emph{RETROFIT-CQ} pipeline has been designed to 
explore the feasibility of retrofitting CQs based on the ontology vocabulary and structure.
It does this through a number of stages: the ontology is parsed to identify and extract the triples representing statements in the ontology; these are then used to generate prompts that are fed into an LLM in order to generate the candidate CQs.
We investigate whether different zero-shot prompts\footnote{We use zero-shot to refer to  prompts expressed without the addition of ground truth examples or fine-tuning.} and the use of different LLMs have an effect on the CQs generated. Figure~\ref{fig:approach} illustrates the main steps in the ~\textit{RETROFIT-CQs} pipeline: 1) triple extraction; 2) question generation; and 3) question filtration. Each of these steps are detailed in the following subsections.

\subsection{Extracting Triples from Ontologies}\label{sec:TripleExtraction}
This task parses an ontology to generate \texttt{statements} in the form of RDF triples, from which we extract the triple components \textit{(s,p,o)} representing the \texttt{`Subject'}, \texttt{`Predicate'}, and \texttt{`Object'} of a statement respectively. In this preliminary study we assume that resources in the triples are represented by HTTP URIs and that have readable local names, i.e. meaningful for human readers~\cite{10.1145/3382097}. Ontologies with opaque local names~\cite{10.1145/3382097} (e.g. Wikidata Q-items) are excluded as well.

Likewise, we exclude from the list of generated triples those that have blank node identifiers as subject or object in the triple, since they would require dedicated processing.\footnote{A version of this paper with a more comprehensive evaluation is in preparation.}

\subsection{LLM-based Question Generation}\label{sec:QuestionGeneration}
In this step we generate the set of prompts that instruct an LLM to generate queries regarding the list of statements (triples) produced (in Section \ref{sec:TripleExtraction}),
as illustrated in Algorithm~\ref{alg:pipeline}. For the purpose of this investigation, we consider different prompt instructions: a general one (Prompt 1) and two with added context (Prompt 2 and 3).\footnote{The precise formulation of the three prompts can be found in Section~\ref{sec:experiment1}.}


The rationale for using entire statements, rather than individual triple components
separately is to provide a contextual boundary for the LLM, therefore limiting the generation of out of scope questions. For example, consider a Solar System ontology that contains the triple \texttt{[`Hippocamp', `type', `Solar\_System\_Satellite']}. For an LLM (e.g. Chat GPT3.5) the term `Hippocamp' can equally refer to the `Hippocampus' in brain anatomy as well as to the `Hippocamp' moon in astronomy.\footnote{Hippocamp is one of the smaller moons of the planet Neptune, discovered in 2013.} Hence, it will generate questions that refer to both these meanings, e.g.  \textit{`What is the primary function of the Hippocamp?'} and \textit{`Is Hippocamp a satellite of Neptune?'}. By using the entire statement we 
provide a form of disambiguation.

\begin{algorithm}[t]
\caption{Generate Questions from Multiple LLMs for Each Prompt}
\label{alg:pipeline}
\begin{algorithmic}[1]
\Require $prompts = [prompt1, prompt2, prompt3]$
\Require $LLMs = [LLM1, LLM2, LLM3]$

\Function{generate\_questions}{$statement$, $prompt$, $LLM$}
    \State $questions \gets []$

    \State \begin{varwidth}[t]{\linewidth}
      $instructions$~$\gets$\par
        \hskip\algorithmicindent \text{Concatenate } $prompt$ \text{ with elements of } $statement$
      \end{varwidth}
    \State $response \gets \text{Call } LLM \text{ with } prompt = instructions$
    \State $question \gets \text{Extract and Clean Text from } response$
    \State \text{Append } $question$ \text{ to } $questions$
    \State \Return $questions$
\EndFunction

\For{each $prompt$ in $prompts$}
    \For{each $LLM$ in $LLMs$}

        \State \begin{varwidth}[t]{\linewidth}
            $questions$~$\gets$~\Call{generate\_questions}{\par
                \hskip\algorithmicindent ["Subject", "Predicate", "Object"], \par
                \hskip\algorithmicindent prompt, LLM}
         \end{varwidth}

        \State $filename \gets \text{"questions\_"} + prompt + \text{"\_"} + LLM + \text{".csv"}$
        \State \Call{save\_to\_csv}{questions, filename}
    \EndFor
\EndFor
\end{algorithmic}
\end{algorithm}

\subsection{Question Filtration} 
In this step we eliminate redundant questions by detecting duplicate ones, and other questions that refer to the modelling primitive used, such as \textit{``Is Multiplayer a class?''}. These questions should be excluded as CQs are designed to be independent of the chosen modelling style. 
In this step we also eliminate questions that require some subjective assessment or that require text generation, e.g. \textit{``Could you envision a future where multiplayer games abandon traditional achievements in favour of more dynamic, player-driven goals and objectives? Why or why not?''}. The primary aim of CQs is to scope the ontology, and provide context in terms of \emph{how, where, when, why, who}~\cite{juan2019}; therefore questions that require the generation of a narrative are not suitable CQs.
At the end of this step, we have a set of well formed generated CQs.


\subsection{Implementation of the Method}
The \textit{RETROFIT-CQ} pipeline is implemented in Python 3.10.12, and is available in a GitHub repository\footnote{\url{https://github.com/SemTech23/RETROFIT-CQs}} as supplementary material
. RDFLib 7.0.0\footnote{\url{https://github.com/RDFLib/rdflib}} is used to process the ontologies and the RDF statements, and FuzzyWuzzy 0.18.0\footnote{\url{https://pypi.org/project/fuzzywuzzy/}} is used to  detect and remove duplicate questions by using approximate string matching.

In order to 
mitigate the bias on the questions generated by the choice of a specific LLM, we run our experiments using three LLM systems, with the following configurations:
\begin{enumerate}
\item \textit{gpt-3.5-turbo model API}: released by OpenAI. We use the pre-trained model with the maximum requested token value set to 4,096 tokens.  Note that this is the current maximum value for the \textit{gpt-3.5-turbo} model, with 1 token approximately corresponding to 4 chars of English text.\footnote{For detailed token calculation refer to: \url{https://help.openai.com/en/articles/4936856-what-are-tokens-and-how-to-count-them}}

\item \textit{gpt-4 model API}: released by OpenAI. We use the pre-trained model with the maximum requested token value set to 8,192 tokens (current maximum value for the \textit{gpt-4} model).
\item \textit{
Llama 2 (Llama-2-70b-chat) model API}: released by Meta AI, with the pre-trained model.\footnote{\url{https://huggingface.co/meta-llama/Llama-2-70b-chat}}
\end{enumerate}

\section{Empirical Evaluation}
\label{sec:Evaluation}
To evaluate the \emph{RETROFIT-CQ} pipeline, two experiments were conducted: the first quantitatively 
matches the CQs generated with our pipeline against the corresponding CQs published for each ontology considered in our study; whereas the second 
solicits a qualitative assessment of the generated CQs from the ontology developer. We need to emphasise that the main aim of these experiments is to explore the feasibility of the proposed approach, therefore addressing the research question identified in Section~\ref{sec:introduction}, rather than providing a comprehensive evaluation of LLMs ability to generate usable CQs.

The first experiment evaluates the pipeline 
across three existing ontologies and their related CQs.
For this experiment we utilised three ontologies from the CORAL CQ repository~\cite{coral}, one of which has also been used by \citeauthor{wisniewski2019analysis}~\cite{wisniewski2019analysis}:
Video Game~\cite{Parkkila2017AnOF}; VICINITY Core\footnote{\url{http://iot.linkeddata.es/def/core/index-en.html}}; and Dem@care.\footnote{\url{https://demcare.eu/wp-content/uploads/2020/11/D5.1SemanticKnowledgeStructures_andRepresentation.pdf}} Each was selected on the basis that they were produced by different developers and that each had a significant number of published CQs.
%
The characteristics of these ontologies are included in Table \ref{tab:unmatchedCQs}.
%
%
%
In the second experiment, we retrofit CQs to an ontology independently developed and for which no explicit CQs were produced.
This was then followed by an interview with the ontology developer to explore, at least anecdotally, the intent of an ontology designer when building an ontology. We selected the Solar System Ontology\footnotemark[7] (with 337 triples), developed within our research group for a different project.

\subsection{Generating Candidate CQs from Repositories of Requirements}\label{sec:experiment1}

\begin{table*}[ht]
\centering
\caption{Candidate CQs for the Triple [`Multiplayer', `subClassOf', `Achievement'] from the Prompts in LLMs}
\label{tab:exampleofCQs}
\renewcommand{\tabcolsep}{2pt}
\small
\begin{tabularx}{\textwidth}{c|c|X}
\hline
\textbf{LLM} & \textbf{Prompt} & \textbf{Candidate CQs} \\
\hline
\hline

\parbox[t]{2mm}{\multirow{11}{*}{\rotatebox[origin=c]{90}{\tt gpt-3.5-turbo}}} & 
  \multirow{3}{*}{\bf P1} & What is a Multiplayer Achievement? \\
    & & What types of Achievements can be earned in a Multiplayer game? \\
    & & How do Multiplayer Achievements compare to Single Player Achievements? \\
\cline{2-3}
  &
  \multirow{4}{*}{\bf P2} & What strategies do you use to succeed in multiplayer games? \\
    & & What techniques do you use to maximize your achievements in multiplayer games? \\
    & & How do you measure your progress in multiplayer games? \\
    & & What do you do to stay ahead of the competition in multiplayer games? \\
\cline{2-3}
  &
  \multirow{4}{*}{\bf P3} & What is the definition of a Multiplayer Achievement? \\
    & & How does a Multiplayer Achievement differ from a single-player Achievement? \\
    & & What strategies can be used to successfully complete a Multiplayer Achievement? \\
    & & How can Multiplayer Achievements be tracked and monitored? \\
\hline
\hline

\parbox[t]{2mm}{\multirow{11}{*}{\rotatebox[origin=c]{90}{\tt gpt-4}}} & 
  \multirow{4}{*}{\bf P1}& How is the multiplayer mode categorized under achievements? \\
    & & What are the achievements specific to the multiplayer category? \\
    & & Does every multiplayer game have its own set of achievements? \\
    & & What is the relationship between a "multiplayer" and an "achievement" in this context? \\
\cline{2-3}
  &
  \multirow{3}{*}{\texttt{\bf P2}} & What class does Multiplayer belong to? \\
    & & Does the Achievement have any subclasses? \\
    & & What is the relationship between Multiplayer and Achievement? \\
\cline{2-3}
  &
  \multirow{4}{*}{\bf P3} & What is the subclass of Multiplayer? \\
    & & Which concepts are subclass of Achievement? \\
    & & Is there a hierarchical relationship between Multiplayer and Achievement? \\
    & & What specific class is under the Achievement class in the ontology? \\
\hline
\hline

\parbox[t]{2mm}{\multirow{10}{*}{\rotatebox[origin=c]{90}{\tt Llama-2-70b-chat}}} & 
  \multirow{2}{*}{\bf P1} & How do players typically earn this achievement in the game? Are there specific requirements or challenges that must be completed? \\
    & & Does the Subclass achievement grant any special rewards or benefits to players who earn it? \\
\cline{2-3}
  &
  \multirow{5}{*}{\bf P2} & What are some common multiplayer game modes that can be considered subclasses of achievements? \\
& & How do multiplayer games typically incorporate achievements into their gameplay mechanics? \\
& & Can you name an example of a multiplayer game that has a strong focus on achievements and describe how they are used in the game? \\
& & How do achievements in multiplayer games differ from those in single-player games? \\
& & In what ways can multiplayer games use achievements to encourage player engagement and competition? \\
\cline{2-3}
  &
  \multirow{3}{*}{\bf P3} & How do different multiplayer game modes influence player behaviour and motivation to achieve rewards? \\
& & Can you design a multiplayer game mode that incorporates achievements as rewards for collaboration and teamwork? \\
& & What are some potential challenges or drawbacks of using achievements in multiplayer games, and how could they be addressed? \\
\hline

\end{tabularx}
\end{table*}

The aim of this exploratory experiment is to assess the feasibility of the proposed approach, therefore addressing our research question, \textbf{Q1: To what extent it is possible to generate ``usable''  CQs from existing ontologies that are representative of the scope and tasks for which the ontology was designed?} 
In the experiment we exploit LLMs to automatically generate questions in natural language, a task for which LLMs have been successfully employed~\cite{DBLP:journals/pai/MullaG23}.
%
%
We use three different formulations of the prompt, adding further contextual information, to assess whether it would affect the questions generated.  This allows us to address an additional, secondary research question:
\textbf{Q2: To what extent does the addition of specific context result in a more accurate generation of CQs?}

After having extracted the triples from each of these three ontologies, we use the selected LLMs, \textit{gpt-3.5-turbo}, \textit{gpt-4}, and \textit{Llama-2-70b-chat}, to generate the questions from three prompt types:
\begin{itemize}[noitemsep,topsep=4pt]
    \item[P1 --] \textit{General Questions}: this prompt instructs an LLM to generate questions for a given statement:~\textit{[``Based on <\textit{statement}>, generate a list of relevant question''+ \textit{statement}.]}. 
    \item[P2 --] \textit{Competency Questions:} this prompt instructs the LLM to explicitly generate CQs for a given statement by explicitly providing the definition of CQs:~\textit{[''Based on the <\textit{statement}>, generate a list of competency question. Definition of competency questions: the questions that outline the scope of an ontology and provide an idea about the knowledge that needs to be entailed in the ontology.''+ \textit{statement}.]}.
    \item[P3 --] \textit{Use of a Role with Competency Questions:} this prompt contextualises the prompt by specifying the role of ``Ontology Engineer'' and instructs the LLM to explicitly generate CQs for a given statement by including the definition of CQs: ~\textit{[``As an ontology engineer, generate a list of competency questions based on the <\textit{statement}>. Definition of competency questions: the questions that outline the scope of ontology and provide an idea about the knowledge that needs to be entailed in the ontology''+ \textit{statement}.]}.
\end{itemize}


\begin{table*}[ht]
\begin{small}
\centering

\begin{tabular}{l|c|l|c|c|c|c|c|c|c}
\cline{2-10}
&\multirow{2}{*}{\textbf{Prompt}} & \multirow{2}{*}{\textbf{LLMs}} & \multirow{2}{*}{\textbf{No. Q.}} & \multirow{2}{*}{\textbf{Mean Q/T}} & \multirow{2}{*}{\textbf{No.  Candidate CQs}} & \multirow{2}{*}{\textbf{No. Validated CQs}} & \multicolumn{3}{c}{\textbf{Performance Metrics}} \\
\cline{8-10}
 & &  &  &  &  &  & \textbf{Precision} & \textbf{Recall} & \textbf{F1 Score} \\
\hline
\hline

\parbox[t]{2mm}{\multirow{9}{*}{\rotatebox[origin=c]{90}{\bf Video Game}}}
  & \multirow{3}{*}{P1} & \texttt{gpt-3.5-turbo} & 549 & 1.51 & 375 & 204 & 0.5440& 0.9622 & 0.6950 \\
    & & \texttt{gpt-4} & 1373 & 3.79 & 1306 & 1115 & 0.8537 & 0.9955 & 0.9192 \\
    & & \texttt{Llama-2-70b-chat} & 497 & 1.37 & 439 & 399 & 0.9088 & 0.9925 & 0.9488 \\
\cline{2-10}
 & \multirow{3}{*}{P2} & \texttt{gpt-3.5-turbo} & 1129 & 3.11 & 902 & 532 & 0.5898 & 0.9943 & 0.7404 \\
 & & \texttt{gpt-4} & 4536 & 12.49 & 1591 & 1406 & 0.8837 & 0.9985 & 0.9376 \\
 & & \texttt{Llama-2-70b-chat} & 2456 & 6.76 & 1050 & 917 & 0.8733& 0.9989 & 0.9319 \\
\cline{2-10}
 & \multirow{3}{*}{P3} & \texttt{gpt-3.5-turbo} & 1078 & 2.97 & 812 & 487 & 0.5997 & 0.9898 & 0.7469 \\
 & & \texttt{gpt-4} & 4180 & 11.51 & 1738 & 1398 & 0.8043 & 0.9971 & 0.8904 \\
 & & \texttt{Llama-2-70b-chat} & 2323 & 6.39 & 1060 & 796 & 0.7509 & 0.9962 & 0.8563 \\

\hline
\hline

\parbox[t]{2mm}{\multirow{9}{*}{\rotatebox[origin=c]{90}{\bf VICINITY Core}}}
 & \multirow{3}{*}{P1} & \texttt{gpt-3.5-turbo} & 620 & 0.71 & 538 & 280 & 0.5204 & 0.9722 & 0.6779 \\
 & & \texttt{gpt-4} & 830 & 0.95 & 565 & 453 & 0.8017 & 0.9956 & 0.8882 \\
 & & \texttt{Llama-2-70b-chat} & 560 & 0.64 & 225 & 135 & 0.6000 & 0.9712 & 0.7417 \\
\cline{2-10}
 & \multirow{3}{*}{P2} & \texttt{gpt-3.5-turbo} & 900 & 1.03 & 690 & 273 & 0.3956 & 0.9545 & 0.5594 \\
 & & \texttt{gpt-4} & 2926 & 3.35 & 1109 & 710 & 0.6402 & 0.9847 & 0.7759 \\
 & & \texttt{Llama-2-70b-chat} & 1843 & 2.11 & 680 & 380 & 0.5588 & 0.9819 & 0.7122 \\
\cline{2-10}
 & \multirow{3}{*}{P3} & \texttt{gpt-3.5-turbo} & 934 & 1.06 & 512 & 236 & 0.4609 & 0.9593 & 0.6226 \\
 & & \texttt{gpt-4} & 1726 & 1.98 & 778 & 428 & 0.5501 & 0.9861 & 0.7062 \\
 & & \texttt{Llama-2-70b-chat} & 2269 & 2.59 & 698 & 448 & 0.6418 & 0.9911 & 0.7791 \\

\hline
\hline

\parbox[t]{2mm}{\multirow{9}{*}{\rotatebox[origin=c]{90}{\bf Dem@care}}}
 & \multirow{3}{*}{P1} & \texttt{gpt-3.5-turbo} & 1900 & 0.84 & 1485 & 998 & 0.6720 & 0.9920 & 0.8012 \\
 & & \texttt{gpt-4} & 4289 & 1.92 & 2789 & 2336 & 0.8348 & 0.9961 & 0.9084 \\
 & & \texttt{Llama-2-70b-chat} & 2453 & 1.09 & 1566 & 1246 & 0.7956 & 0.9928 & 0.8833 \\
\cline{2-10}
 & \multirow{3}{*}{P2} & \texttt{gpt-3.5-turbo} & 3439 & 1.54 & 1372 & 1006 & 0.7332 & 0.9940 & 0.8439 \\
 & & \texttt{gpt-4} & 4159 & 1.86 & 2342 & 2022 & 0.8633 & 0.9970 & 0.9254 \\
 & & \texttt{Llama-2-70b-chat} & 2414& 1.07 & 1500& 1220 & 0.8133 & 0.9934 & 0.8944 \\
\cline{2-10}
 & \multirow{3}{*}{P3} & \texttt{gpt-3.5-turbo} & 3590 & 1.60& 1248& 856 & 0.6858 & 0.9907 & 0.8106 \\
 & & \texttt{gpt-4} & 4843 & 2.16 & 2874 & 2580 & 0.8977 & 0.9984 & 0.9454 \\
 & & \texttt{Llama-2-70b-chat} & 2413& 1.08 & 1586& 1321& 0.8329 & 0.9947 & 0.9066 \\

\hline

\end{tabular}

\caption{Summary for each prompt in the LLMs: number of generated questions (No. Q.),  mean questions per triple (Mean Q/T), filtered questions in the final output (No. Candidate CQs), number of validated
candidate CQs against existing CQs (No. of Validated CQs) and Performance Metrics including Precision, Recall \& F1 score}
\label{tab:summary_video_game_and_core}
\end{small}
\end{table*}

The prompts generate questions for each extracted statement of type \textit{(s,p,o)},
where we filter out statements whose subject or object are blank nodes, as discussed in Section~\ref{sec:approach}.
Table \ref{tab:exampleofCQs} illustrates some of the questions generated by each prompt for the triple \textit{Multiplayer, subClassOf, Achievement)}  from the Video Game ontology. The candidate CQs, resulting from the question filtration step,
are validated to check whether they match the design-stage CQs reported for these ontologies. 
In order to mitigate the effect of paraphrasing, or the use of different morphological structures (e.g. plurals) on the similarity assessment, we employed \texttt{SBERT}, which is a variant of the pretrained BERT approach that derives semantically meaningful sentence embeddings from siamese and triplet network structures, as it can be used for semantic similarity and paraphrase detection~\cite{reimers-gurevych-2019-sentence}. 
The results are summarised in
Table~\ref{tab:summary_video_game_and_core}, where for each ontology we report the following performance metrics for each prompt and LLM: number of generated questions (No. of Q.), mean questions per triple (Mean Q/T), filtered questions, i.e. the final output (No. Candidate CQs), number of validated candidate CQs against existing CQs (No. Validated CQs), and traditional precision, recall and F-measure values. 

\subsection{Retrofitting CQ Results}
\label{sec:coral-results}

The results obtained by retrofitting CQs to the Video Game, VICINITY Core and Dem@care ontologies show that all models generate a significant number of candidate CQs, as evidenced by the high recall scores. Our approach achieves a recall of 0.95 or above for all prompts and LLMs, with the three lowest scores recorded in the Vicinity Core Ontology for \texttt{gpt-3.5-turbo} with Prompt 2 (0.95), 3 (0.95) and 1 (0.97), respectively. Therefore, we match the majority of the design CQs catalogued in CORAL for the three ontologies, and we confirm the usability of our generated CQs as they accurately identify the design CQs.
The results, however, are not as clear when we consider precision. There are a number of design CQs that are not matched, as well as CQs that are not relevant, as can be seen in the column ``\textbf{(UnmatchedCQs) \%}'' in Table~\ref{tab:unmatchedCQs}. The worse overall precision is recorded for VICINITY Core, where precision varies between 0.39 and 0.80. For Prompt 3, which is where we provided the role of the ontology developer and the definition of CQs, all of the LLMs record lower precision: 0.46 for \texttt{gpt-3.5-turbo}, 0.55 for \texttt{gpt-4} and 0.64 for \texttt{Llama-2-70b-chat}. 

Regarding the choice of different LLMs, \texttt{gpt-3.5-turbo} yields the lowest precision for each of the three ontologies and prompts, except in the case of Prompt 3 when applied to the triples extracted from Dem@care, which achieves the best precision across the models. 
Adding contextual information to the prompt seems to yield a limited improvement in precision and in some cases no improvement at all (as in the case of VICINITY Core, where the highest precision for \texttt{gpt-3.5-turbo} is obtained when executing Prompt 1, and the precision for Prompt 3 over the three LLMs is almost always worse than the precision for Prompt 1 and 2).

\begin{table*}[ht]
\small
\centering
\caption{Descriptive statistics for unmatched CQs from RETROFIT-CQs based on different prompts: ontology name, LLMs, unmatched CQ count and unmatched percentage ((UnmatchedCQs) \%), Mean, Standard Deviation (Std), word count range (Min, 0.25, 0.50, Max). The Std value of 0 occurs when the CQs have the same number of words and value, whereas (-) indicates that the value is not available due to there being only one unmatched CQ.}
\label{tab:unmatchedCQs}

\begin{tabular}{l|c|l|c|c|c|c|c|c|c}
\cline{2-10}
 & \textbf{Prompt}  & \textbf{LLMs} & \textbf{(UnmatchedCQs) \%} & \textbf{Mean} & \textbf{Std} & \textbf{Min} & \textbf{0.25} & \textbf{0.50} & \textbf{Max} \\
\hline
\hline

\parbox[t]{4mm}{\multirow{9}{*}{\rotatebox[origin=c]{90}{\bf Video Game}}}
\parbox[t]{2.5mm}{\multirow{9}{*}{\rotatebox[origin=c]{90}{No. of Triples: 363}}}
\parbox[t]{2.5mm}{\multirow{9}{*}{\rotatebox[origin=c]{90}{No. of Design CQs: 66}}}

  & \multirow{3}{*}{P1} 
    & \texttt{gpt-3.5-turbo} & (8) 12\% & 9.62 & 3.02& 7& 7.75 & 9 & 16 \\
    \cline{3-10}
     & & \texttt{gpt-4 }& (5) 8\% & 8.6 & 0.55 & 8 &8 &9 & 9 \\
    \cline{3-10}
     & & \texttt{Llama-2-70b-chat}& (3) 5\% &8.67 & 0.57 & 8 & 8.5 &9 &9 \\
    \cline{2-10}
  & \multirow{3}{*}{P2} 
    & \texttt{gpt-3.5-turbo} & (3) 5\% & 8.33& 0.58 & 8& 8 & 8 & 9 \\
    \cline{3-10}
    & & \texttt{gpt-4} & (2) 3\% &11.5& 4.95 & 8 & 9.75 &11.5& 15 \\
    \cline{3-10}
    & & \texttt{Llama-2-70b-chat}& (1) 2\% & 9 &- & 9 & 9 & 9 & 9 \\
    \cline{2-10}
  & \multirow{3}{*}{P3} 
    & \texttt{gpt-3.5-turbo} & (5) 8\% & 9.6& 3.05 & 8& 8 & 8 & 15 \\
    \cline{3-10}
    & & \texttt{gpt-4} & (4) 6\% &10& 3.37 & 8 & 8 &8.5& 15 \\
    \cline{3-10}
    & & \texttt{Llama-2-70b-chat}& (3) 5\% & 8.33 &1.15 & 7 & 8 & 9 & 9 \\

\hline
\hline

\parbox[t]{4mm}{\multirow{9}{*}{\rotatebox[origin=c]{90}{\bf VICINITY Core}}}
\parbox[t]{2.5mm}{\multirow{9}{*}{\rotatebox[origin=c]{90}{No. of Triples: 873}}}
\parbox[t]{2.5mm}{\multirow{9}{*}{\rotatebox[origin=c]{90}{No. of Design CQs: 57}}}

  & \multirow{3}{*}{P1} 
    & \texttt{gpt-3.5-turbo} & (8) 14\% &7.62 & 2.66 & 4 & 5.5 & 8.5 &11 \\
    \cline{3-10}
    & & \texttt{gpt-4} & (2) 4\% &4 &0 & 4 & 4& 4 &4 \\
    \cline{3-10}
    & & \texttt{Llama-2-70b-chat} & (4) 7\% &5.5&  1.91 & 4 & 4 &5 & 8 \\
    \cline{2-10}
  & \multirow{3}{*}{P2} 
    & \texttt{gpt-3.5-turbo} & (13) 23\% & 6.54 & 2.03 & 4 & 4 & 7 & 10 \\
    \cline{3-10}
    & & \texttt{gpt-4} & (11) 19\% &6.73 & 1.74 & 4 & 6& 7 & 10 \\
    \cline{3-10}
    & & \texttt{Llama-2-70b-chat} & (7) 12\% &6&  1.53 &4 & 5 &6 & 8 \\
    \cline{2-10}
  & \multirow{3}{*}{P3} 
    & \texttt{gpt-3.5-turbo} & (10) 18\% & 6.5 & 2.51 & 4 & 4 & 6.5 & 11 \\
    \cline{3-10}
    & & \texttt{gpt-4} & (6) 11\% &6 & 1.67 & 4 & 4.5&6.5 & 8 \\
    \cline{3-10}
    & & \texttt{Llama-2-70b-chat} & (4) 7\% &5.25&  1.5 &4 & 4 &5 & 7 \\

\hline
\hline

\parbox[t]{4mm}{\multirow{9}{*}{\rotatebox[origin=c]{90}{\bf Dem@care}}}
\parbox[t]{2.5mm}{\multirow{9}{*}{\rotatebox[origin=c]{90}{No. of Triples: 2238}}}
\parbox[t]{2.5mm}{\multirow{9}{*}{\rotatebox[origin=c]{90}{No. of Design CQs: 107}}}

  & \multirow{3}{*}{P1} 
    & \texttt{gpt-3.5-turbo} & (8) 7\% & 7.63 & 1.51& 5 & 7 & 7.5 & 10\\
    \cline{3-10}
    & & \texttt{gpt-4} & (9) 8\% &7.56 &1.88 & 4 & 7&8 & 9\\
    \cline{3-10}
    & & \texttt{Llama-2-70b-chat} & (9) 8\% &7.78&  2.86 &4 & 7 &7 & 14\\
    \cline{2-10}
  & \multirow{3}{*}{P2} 
    & \texttt{gpt-3.5-turbo} & (6) 6\% & 8.5 & 3.02& 5 & 7.25 & 8 & 14\\
    \cline{3-10}
    & & \texttt{gpt-4} & (6) 6\% &8.17 &1.17 & 7 & 7.25&8 & 10\\
    \cline{3-10}
    & & \texttt{Llama-2-70b-chat} & (8) 7\% &6.63&  1.6 &4 & 5.75 &7 & 9\\
    \cline{2-10}
  & \multirow{3}{*}{P3} 
    & \texttt{gpt-3.5-turbo} & (8) 7\% & 7.5 & 0.76 & 7 & 7 & 7.5 & 9 \\
    \cline{3-10}
    & & \texttt{gpt-4} & (4) 4\% &8 &0.82 & 7 & 7.75&8 & 9 \\
    \cline{3-10}
    & & \texttt{Llama-2-70b-chat} & (7) 7\% &6.86&  1.77 &4 & 6 &7 & 9\\

\hline

\end{tabular}
\end{table*}

Some of these variations can be explained by the way the LLMs formulate questions about classes and their properties, especially for Prompts 2 and 3 where we specify the definition of CQs. In these cases, the LLMs formulate questions by asking about the property and its relation to the class, or conversely, about the class and its relation to the property. For example, 
for the triple where \textit{username} is a datatype property whose domain is the class \textit{Player}  in the Video Game ontology,
the LLMs generate the following questions for Prompt 3:
\begin{itemize}[noitemsep,topsep=0pt]
\item 
\textit{What is the relationship between the username and the player?}
\item 
\textit{Is there a player associated with every username?}
\item 
\textit{Does every player have a username?}
\end{itemize}
However, the injection of the definition of CQ and the ontology developer role results in candidate CQs expressed in terms of ontological modelling, and therefore far from what an ontology developer or engineer would write. 
Such a difference in phrasing style can potentially undermine the similarity matching of the CQs.
For example, the question \textit{``How does a username connect a player to a text string?''} generated by Prompt 3 used with \texttt{gpt-4} does not match the design CQ \textit{``What is the username of the player?''}, despite having the same meaning.  

%

Some of the design CQs in the ontologies, however, are missed completely by this process. The reasons for this are discussed in more detail in Section~\ref{sec:discussion}.

\subsection{Ontology Developer-Led Evaluation}

\begin{table*}[ht]
\begin{small}
\centering
\begin{tabular}{l|c|c|c|c|c}
\hline
\textbf{LLMs} & \textbf{No. of Q.} & \textbf{Mean Q/T} & \textbf{No. of candidate CQs} & \textbf{No. of validated CQs} & \textbf{Precision} \\
\hline
\texttt{gpt-3.5-turbo} & 604 & 1.79 & 206 & 180 & 0.8737 \\
\texttt{gpt-4} & 1194 & 3.54 & 800 & 609 & 0.7612 \\
\texttt{Llama-2-70b-chat} & 1856 &5.51 & 1108 & 995 & 0.9074 \\
\hline
\end{tabular}
\end{small}
\caption{Solar System Ontology -- For each LLM using Prompt 1 we report: number of generated questions (No. of Q.), mean questions per triple (Mean Q/T), filtered questions (No. of Candidate CQs), developer-validated candidate CQs (No. of Validated CQs), and the Precision.}
\label{tab:solar_system}
\end{table*}

The results obtained from the numerical evaluation presented in the previous section provides several useful insights on this exploratory study. However, the assessment of similarity between the design CQs and the candidate CQs cannot fully take into account the aims and intentions of the ontology engineer when writing the CQs. We therefore applied the \textit{RETROFIT-CQ} pipeline to the Solar System Ontology and asked the ontology developer~\footnote{We would like to thank Samah Alkhuzaey for her contribution and assistance with the experiments.} to evaluate the candidate CQs. The Solar System Ontology represents astronomical knowledge 
that is intended to support the generation of Multiple Choice Questions (MCQs) for exam papers aimed at secondary school students \cite{AlkhuzaeyECEL23}. We interviewed the ontology developer to assess the candidate CQs to identify: 1) correct candidate CQs, and 2) candidate CQs that did not match the design CQs but that in hindsight could have been considered valid for the purpose and scope of the ontology.

The methodology for this evaluation is similar to that used in the previous evaluation (and reported in Section \ref{sec:coral-results}); however for this evaluation, we only explore the use of Prompt 1 with the three LLMs.
Our aim is to assess whether, even with the simplest of settings, the LLMs generate questions that an ontology engineer would consider usable and that reflect their design aims.
Table~\ref{tab:solar_system} reports the result of applying \textit{RETROFIT-CQ} to the Solar System Ontology,
and for each LLM, lists: the number of generated questions (No. of Q.); mean questions per triple (Mean Q/T); the number of candidate CQs (No. of Candidate CQs) which also corresponds to the number of filtered questions;
the number of validated CQs (No. of Validated CQs) representing the number of candidate CQs that were validated by the developer; and the precision (based on the ratio of validated CQs to the number of candidate CQs).

The interview conducted with the ontology developer confirmed that the CQs generated by the pipeline accurately captured the initial requirements (with a precision over 0.75) and that all the design CQs were matched. Interestingly, the developer identified a number of additional candidate CQs that in hindsight captured some of the intended meaning of the ontology and could have been included in the initial requirement elicitation phase. Some of these CQs refer to specific named individuals e.g. ``How long has Tethys been in operation?'' and were formulated differently.
Others, however, were not included in the design CQs but capture the knowledge modelled in the ontology; e.g. ``What are the technical criteria required for a celestial body to be classified as a planet?'', or ``What is composed of silicate rocks or metals?'' 
Although ontology developers write CQs that are representative of the requirements of the ontology, there is no guarantee that the list produced is exhaustive or comprehensive, and thus by retrofitting CQs from the ontology, additional CQs can emerge. 
This additional set of CQs 
could also be useful when evaluating the ontology design by: 1) translating CQs into SPARQL queries; and 2) query the populated ontology to stress test the ontology design process and anticipate unintended uses of the ontology.

\section{Discussion}\label{sec:discussion}

The motivation for our work is to facilitate the reuse of ontologies that have no associated available CQs,  by \emph{retrofitting} a viable set of candidate CQs for such ontologies that reflect the original aims of the ontology engineer~\cite{Alharbi2021AssessingCO,Aziz2023}.
%
%
The results from the exploratory experiments described in Section~\ref{sec:Evaluation} are encouraging and indicate the viability of using generative approaches to retrofit competency questions from existing ontologies when their design CQs are not available. We use the LLMs in their default configurations, as the focus of this exploratory study is the CQ generation, rather than an analysis of the performance of different LLMs. Even when using a combination of: 1) default settings, 2) prompts that do not use any examples or external KBs for inferring the related information, and 3) a pipeline that is agnostic wrt the choice of prompt or LLM, we can still achieve recall values that are very close to 1. However, LLMs are very effective in generating natural language questions that are amenable to being paraphrased and reformulated, which is one of the reasons why we see such a large variation in precision values (ranging from 0.3956 to 0.9088) which seems to suggest that limiting the creativity of the LLMs can bring results that are more deterministic and produce fewer questions. In general, adding contextual information in the form of the definition of CQs seems to improve precision; however, with some LLMs it also increases the complexity and length of the questions that are framed in terms of ontological modelling, and thus may be less similar to the types of CQs written by practitioners (as discussed in Section~\ref{sec:Evaluation}). One positive consequence of the \emph{RETROFIT-CQ} approach is that the resulting CQs provide a snapshot of how the intended model has been represented in the ontology and can detect any unintended modelling consequences, as highlighted by the results of the ontology developer based validation.

Furthermore, there are a small number of design CQs that are not matched, as shown in Table~\ref{tab:unmatchedCQs}, which warrants a closer investigation. An analysis of the design questions that are unmatched reveals that these fall under (a combination of) 5 main categories: 1) questions requiring calculation or some aggregation function, e.g. \textit{``Who are the top 3 players in the game?''}; 2) multi-hop questions (spanning more than one triple), e.g. \textit{``What functional areas are of clinical relevance for the home and nursing home environments?''}  3) CQs with no corresponding or ambiguous ontology content, e.g. \textit{``Which devices are located at UNIKL?''};
4)  CQs that are poorly phrased e.g. \textit{``Which properties from a panic button observed in events?''}; and finally, 5) CQs whose answer involves some level of subjectivity from the developers perspective,
e.g.
\textit{``Which devices can I see?''}. 

Questions from category 1 require some level of calculation over the information modelled in the ontology, and LLMs have well documented limitations when reasoning over mathematical notions~\cite{DBLP:conf/acl/ImaniD023}, which can lead to the generation of incorrect candidate CQs.  Categories 3, 4, and 5 are all caused by the lack of consistent guidelines for writing competency questions~\cite{wisniewski2019analysis}, and display some of the problems identified~\cite{Ren2014}. For example, the CQ given above in category 3 is not matched because the ontology does not contain a triple whose object is \textit{UNIKL} (i.e. it does not contain any class or named individual whose label is \textit{UNIKL}) and hence it violates one of the presuppositions in \cite{Ren2014} that states that
any element mentioned in a CQ should occur in the ontology (\emph{Presupposition Question Type}). Some CQs are phrased in poor English, therefore increasing the likelihood that they are not matched by any question generated by LLMs, which have been effectively used for Automatic Question Generation~\cite{DBLP:journals/pai/MullaG23}. In other cases, the labels that are used to refer to the ontology elements can lead to confusion. For example, the Video Game ontology contains the triple \textit{(Multiplayer subClassOf Achievement)}. If the triple is taken in isolation (e.g. when an ontology developer who wants to reuse an ontology reads it without considering the documentation) it can be misleading and this ambiguity is reflected in the questions generated by the LLM as seen in Table 2. However, the documentation clarifies that the class \emph{Multiplayer} models a \emph{Multiplayer Achievement}, which would have been a more appropriate label. This type of issue is particularly troublesome when ontology developers that are seeking to reuse an ontology try to match terms they used in their requirements.

Category 2 questions highlight a different problem. 
Although our pipeline is configured to only generate single hop (as opposed to multi hop) questions as we focus on statements \textit{(s,p,o)}, the results still suggest that we match the majority of design CQs (as the recall is close to 1).  This also suggests that 
those design CQs are also single hop, possibly due to the way knowledge has been modelled.
%
%
However, if we consider examples of enterprise ontologies, whose model is typically an abstraction of one or more database schemas, we see that CQs are typically more complex and often involve more than one entity; e.g. the CQ ``How many orders were placed in a given time period per their status?'' was used as an example in~\cite{juan2019}. Therefore, future work should address the generation of candidate CQs from several triples at the same time.

\section{Conclusions} \label{sec:conclusions}
This paper presents an exploratory study in retrofitting CQs on existing ontologies. It proposes a pipeline that exploits LLMs in order to automatically generate natural language questions about each triple in the ontology. In this study, we considered 3 LLMs, \texttt{gpt-3.5-turbo, gtp-4 and Llama-2-70b-chat}, and we investigate the use of different prompts, where: Prompt 1 asks to generate questions about a triple; Prompt 2 also adds the definition of CQ; and Prompt 3
extends this further by adding the role ontology engineer. We evaluate the pipeline over three ontologies and their respective competency questions, and observe that our approach has a recall close to 1, i.e. the CQs generated match the design CQs but with varying precision. We have investigated the reasons for the variations in precision, and we conducted an experiment with ontology developers, who were asked to assess the veracity of the generated CQs. 
Furthermore, we analyse the potential reasons for the observed performance. As future work, we plan a more comprehensive evaluation to confirm the findings of this paper wrt a larger corpus of ontologies and CQs.
We will also investigate how adding further tasks, such as paraphrasing, translating, and comparing can fully exploit the
LLMs' language capabilities
to produce a more comprehensive set of prompt templates.
\balance 
\bibliographystyle{ACM-Reference-Format}
\bibliography{main}


\begin{thebibliography}{28}


\ifx \showCODEN    \undefined \def \showCODEN     #1{\unskip}     \fi
\ifx \showDOI      \undefined \def \showDOI       #1{#1}\fi
\ifx \showISBNx    \undefined \def \showISBNx     #1{\unskip}     \fi
\ifx \showISBNxiii \undefined \def \showISBNxiii  #1{\unskip}     \fi
\ifx \showISSN     \undefined \def \showISSN      #1{\unskip}     \fi
\ifx \showLCCN     \undefined \def \showLCCN      #1{\unskip}     \fi
\ifx \shownote     \undefined \def \shownote      #1{#1}          \fi
\ifx \showarticletitle \undefined \def \showarticletitle #1{#1}   \fi
\ifx \showURL      \undefined \def \showURL       {\relax}        \fi
\providecommand\bibfield[2]{#2}
\providecommand\bibinfo[2]{#2}
\providecommand\natexlab[1]{#1}
\providecommand\showeprint[2][]{arXiv:#2}

\bibitem[\protect\citeauthoryear{Alharbi}{Alharbi}{2021}]%
        {Alharbi2021AssessingCO}
\bibfield{author}{\bibinfo{person}{Reham Alharbi}.} \bibinfo{year}{2021}\natexlab{}.
\newblock \showarticletitle{Assessing Candidate Ontologies for Reuse}. In \bibinfo{booktitle}{\emph{Proceedings of the Doctoral Consortium at ISWC 2021 (ISWC-DC)}}.
\newblock
\urldef\tempurl%
\url{https://api.semanticscholar.org/CorpusID:244895203}
\showURL{%
\tempurl}


\bibitem[\protect\citeauthoryear{Alharbi, Tamma, and Grasso}{Alharbi et~al\mbox{.}}{2021}]%
        {alharbiKcap2021}
\bibfield{author}{\bibinfo{person}{Reham Alharbi}, \bibinfo{person}{Valentina Tamma}, {and} \bibinfo{person}{Floriana Grasso}.} \bibinfo{year}{2021}\natexlab{}.
\newblock \showarticletitle{Characterising the Gap Between Theory and Practice of Ontology Reuse}. In \bibinfo{booktitle}{\emph{Proceedings of the 11th on Knowledge Capture Conference}} \emph{(\bibinfo{series}{K-CAP '21})}. \bibinfo{publisher}{ACM}, \bibinfo{pages}{217–224}.
\newblock


\bibitem[\protect\citeauthoryear{AlKhuzaey, Grasso, Payne, and Tamma}{AlKhuzaey et~al\mbox{.}}{2023}]%
        {AlkhuzaeyECEL23}
\bibfield{author}{\bibinfo{person}{Samah AlKhuzaey}, \bibinfo{person}{Floriana Grasso}, \bibinfo{person}{Terry~R. Payne}, {and} \bibinfo{person}{Valentina Tamma}.} \bibinfo{year}{2023}\natexlab{}.
\newblock \showarticletitle{A Framework for Assessing the Complexity of Auto Generated Questions from Ontologies}. In \bibinfo{booktitle}{\emph{Proc. of the 22nd European Conf. on e-Learning (to appear)}}.
\newblock


\bibitem[\protect\citeauthoryear{Allemang, Hendler, and Gandon}{Allemang et~al\mbox{.}}{2020}]%
        {10.1145/3382097}
\bibfield{author}{\bibinfo{person}{Dean Allemang}, \bibinfo{person}{Jim Hendler}, {and} \bibinfo{person}{Fabien Gandon}.} \bibinfo{year}{2020}\natexlab{}.
\newblock \bibinfo{booktitle}{\emph{Semantic Web for the Working Ontologist: Effective Modeling for Linked Data, RDFS, and OWL} (\bibinfo{edition}{3} ed.)}. Vol.~\bibinfo{volume}{33}.
\newblock \bibinfo{publisher}{Association for Computing Machinery}, \bibinfo{address}{New York, NY, USA}.
\newblock
\showISBNx{9781450376174}


\bibitem[\protect\citeauthoryear{Azzi, Assi, and Gagnon}{Azzi et~al\mbox{.}}{2023}]%
        {Aziz2023}
\bibfield{author}{\bibinfo{person}{Sabrina Azzi}, \bibinfo{person}{Ali Assi}, {and} \bibinfo{person}{St{\'e}phane Gagnon}.} \bibinfo{year}{2023}\natexlab{}.
\newblock \showarticletitle{Scoring Ontologies for Reuse: An Approach for Fitting Semantic Requirements}. In \bibinfo{booktitle}{\emph{Metadata and Semantic Research}}. \bibinfo{publisher}{Springer Nature}, \bibinfo{pages}{203--208}.
\newblock


\bibitem[\protect\citeauthoryear{Bezerra and Freitas}{Bezerra and Freitas}{2017}]%
        {Bezerra2017}
\bibfield{author}{\bibinfo{person}{Camila Bezerra} {and} \bibinfo{person}{Fred Freitas}.} \bibinfo{year}{2017}\natexlab{}.
\newblock \showarticletitle{Verifying Description Logic Ontologies based on Competency Questions and Unit Testing}. In \bibinfo{booktitle}{\emph{Proceedings of the {IX} Seminar on Ontology Research in Brazil and {I} Doctoral and Masters Consortium on Ontologies, Bras{\'{\i}}lia, Brazil, August 28th-30th, 2017}} \emph{(\bibinfo{series}{{CEUR} Workshop Proceedings}, Vol.~\bibinfo{volume}{1908})}. \bibinfo{publisher}{CEUR-WS.org}, \bibinfo{pages}{159--164}.
\newblock


\bibitem[\protect\citeauthoryear{Espinoza-Arias, Garijo, and Corcho}{Espinoza-Arias et~al\mbox{.}}{2022}]%
        {Oscar2022}
\bibfield{author}{\bibinfo{person}{Paola Espinoza-Arias}, \bibinfo{person}{Daniel Garijo}, {and} \bibinfo{person}{Oscar Corcho}.} \bibinfo{year}{2022}\natexlab{}.
\newblock \showarticletitle{Extending Ontology Engineering Practices to Facilitate Application Development}. In \bibinfo{booktitle}{\emph{Knowledge Engineering and Knowledge Management}}. \bibinfo{pages}{19--35}.
\newblock


\bibitem[\protect\citeauthoryear{Fern{\'a}ndez-Izquierdo, Poveda-Villal{\'o}n, and Garc{\'i}a-Castro}{Fern{\'a}ndez-Izquierdo et~al\mbox{.}}{2019}]%
        {coral}
\bibfield{author}{\bibinfo{person}{Alba Fern{\'a}ndez-Izquierdo}, \bibinfo{person}{Mar{\'i}a Poveda-Villal{\'o}n}, {and} \bibinfo{person}{Ra{\'u}l Garc{\'i}a-Castro}.} \bibinfo{year}{2019}\natexlab{}.
\newblock \showarticletitle{CORAL: A Corpus of Ontological Requirements Annotated with Lexico-Syntactic Patterns}. In \bibinfo{booktitle}{\emph{The Semantic Web, 16th International Conference, ESWC 2019}}. \bibinfo{publisher}{Springer International Publishing}, \bibinfo{address}{Cham, Switzerland}, \bibinfo{pages}{443--458}.
\newblock


\bibitem[\protect\citeauthoryear{Gangemi, Lippolis, Lodi, and Nuzzolese}{Gangemi et~al\mbox{.}}{2022}]%
        {DBLP:conf/i-semantics/GangemiLLN22}
\bibfield{author}{\bibinfo{person}{Aldo Gangemi}, \bibinfo{person}{Anna~Sofia Lippolis}, \bibinfo{person}{Giorgia Lodi}, {and} \bibinfo{person}{Andrea~Giovanni Nuzzolese}.} \bibinfo{year}{2022}\natexlab{}.
\newblock \showarticletitle{Automatically Drafting Ontologies from Competency Questions with FrODO}. In \bibinfo{booktitle}{\emph{Towards a Knowledge-Aware {AI} - SEMANTiCS 2022 - Proceedings of the 18th International Conference on Semantic Systems, 13-15 September 2022, Vienna, Austria}} \emph{(\bibinfo{series}{Studies on the Semantic Web}, Vol.~\bibinfo{volume}{55})}. \bibinfo{publisher}{{IOS} Press}, \bibinfo{pages}{107--121}.
\newblock


\bibitem[\protect\citeauthoryear{Gr{\"u}ninger and Fox}{Gr{\"u}ninger and Fox}{1995}]%
        {gruninger1995methodology}
\bibfield{author}{\bibinfo{person}{Michael Gr{\"u}ninger} {and} \bibinfo{person}{Mark~S. Fox}.} \bibinfo{year}{1995}\natexlab{}.
\newblock \showarticletitle{The Role of Competency Questions in Enterprise Engineering}.
\newblock In \bibinfo{booktitle}{\emph{Benchmarking --- Theory and Practice}}. \bibinfo{pages}{22--31}.
\newblock


\bibitem[\protect\citeauthoryear{Imani, Du, and Shrivastava}{Imani et~al\mbox{.}}{2023}]%
        {DBLP:conf/acl/ImaniD023}
\bibfield{author}{\bibinfo{person}{Shima Imani}, \bibinfo{person}{Liang Du}, {and} \bibinfo{person}{Harsh Shrivastava}.} \bibinfo{year}{2023}\natexlab{}.
\newblock \showarticletitle{MathPrompter: Mathematical Reasoning using Large Language Models}. In \bibinfo{booktitle}{\emph{Proceedings of the The 61st Annual Meeting of the Association for Computational Linguistics: Industry Track, {ACL} 2023, Toronto, Canada, July 9-14, 2023}}. \bibinfo{pages}{37--42}.
\newblock


\bibitem[\protect\citeauthoryear{Keet and {\L}awrynowicz}{Keet and {\L}awrynowicz}{2016}]%
        {Keettestdriven2016}
\bibfield{author}{\bibinfo{person}{C.~Maria Keet} {and} \bibinfo{person}{Agnieszka {\L}awrynowicz}.} \bibinfo{year}{2016}\natexlab{}.
\newblock \showarticletitle{Test-Driven Development of Ontologies}. In \bibinfo{booktitle}{\emph{The Semantic Web. Latest Advances and New Domains}}. \bibinfo{pages}{642--657}.
\newblock


\bibitem[\protect\citeauthoryear{Keet, Mahlaza, and Antia}{Keet et~al\mbox{.}}{2019}]%
        {CLARO2019}
\bibfield{author}{\bibinfo{person}{C.~Maria Keet}, \bibinfo{person}{Zola Mahlaza}, {and} \bibinfo{person}{Mary-Jane Antia}.} \bibinfo{year}{2019}\natexlab{}.
\newblock \showarticletitle{CLaRO: A Controlled Language for Authoring Competency Questions}. In \bibinfo{booktitle}{\emph{Metadata and Semantic Research}}. \bibinfo{publisher}{Springer International Publishing}.
\newblock


\bibitem[\protect\citeauthoryear{Liu, Yuan, Fu, Jiang, Hayashi, and Neubig}{Liu et~al\mbox{.}}{2023}]%
        {DBLP:journals/csur/LiuYFJHN23}
\bibfield{author}{\bibinfo{person}{Pengfei Liu}, \bibinfo{person}{Weizhe Yuan}, \bibinfo{person}{Jinlan Fu}, \bibinfo{person}{Zhengbao Jiang}, \bibinfo{person}{Hiroaki Hayashi}, {and} \bibinfo{person}{Graham Neubig}.} \bibinfo{year}{2023}\natexlab{}.
\newblock \showarticletitle{Pre-train, Prompt, and Predict: {A} Systematic Survey of Prompting Methods in Natural Language Processing}.
\newblock \bibinfo{journal}{\emph{{ACM} Comput. Surv.}} \bibinfo{volume}{55}, \bibinfo{number}{9} (\bibinfo{year}{2023}), \bibinfo{pages}{195:1--195:35}.
\newblock


\bibitem[\protect\citeauthoryear{Matentzoglu, Malone, Mungall, and Stevens}{Matentzoglu et~al\mbox{.}}{2018}]%
        {adee4fe404784a1a955f5d15beb11e81}
\bibfield{author}{\bibinfo{person}{Nicolas Matentzoglu}, \bibinfo{person}{James Malone}, \bibinfo{person}{Chris Mungall}, {and} \bibinfo{person}{Robert Stevens}.} \bibinfo{year}{2018}\natexlab{}.
\newblock \showarticletitle{MIRO: Guidelines for Minimum Information for the Reporting of an Ontology}.
\newblock \bibinfo{journal}{\emph{Biomedical Semantics}} \bibinfo{volume}{9}, \bibinfo{number}{6} (\bibinfo{year}{2018}).
\newblock


\bibitem[\protect\citeauthoryear{Mulla and Gharpure}{Mulla and Gharpure}{2023}]%
        {DBLP:journals/pai/MullaG23}
\bibfield{author}{\bibinfo{person}{Nikahat Mulla} {and} \bibinfo{person}{Prachi Gharpure}.} \bibinfo{year}{2023}\natexlab{}.
\newblock \showarticletitle{Automatic question generation: a review of methodologies, datasets, evaluation metrics, and applications}.
\newblock \bibinfo{journal}{\emph{Prog. Artif. Intell.}} \bibinfo{volume}{12}, \bibinfo{number}{1} (\bibinfo{year}{2023}), \bibinfo{pages}{1--32}.
\newblock


\bibitem[\protect\citeauthoryear{Noy, McGuinness, et~al\mbox{.}}{Noy et~al\mbox{.}}{2001}]%
        {noy2001ontology}
\bibfield{author}{\bibinfo{person}{Natalya~F Noy}, \bibinfo{person}{Deborah~L McGuinness}, {et~al\mbox{.}}} \bibinfo{year}{2001}\natexlab{}.
\newblock \bibinfo{booktitle}{\emph{Ontology development 101: A guide to creating your first ontology}}.
\newblock \bibinfo{type}{{T}echnical {R}eport}. \bibinfo{institution}{Stanford knowledge systems laboratory technical report KSL-01-05}.
\newblock


\bibitem[\protect\citeauthoryear{Ouyang et~al\mbox{.}}{Ouyang et~al\mbox{.}}{2022}]%
        {DBLP:conf/nips/Ouyang0JAWMZASR22}
\bibfield{author}{\bibinfo{person}{Long Ouyang} {et~al\mbox{.}}} \bibinfo{year}{2022}\natexlab{}.
\newblock \showarticletitle{Training language models to follow instructions with human feedback}. In \bibinfo{booktitle}{\emph{Advances in Neural Information Processing Systems, NeurIPS 2022}}, Vol.~\bibinfo{volume}{35}. \bibinfo{publisher}{Curran Associates, Inc.}, \bibinfo{pages}{27730--27744}.
\newblock


\bibitem[\protect\citeauthoryear{Parkkila, Radulovic, Garijo, Poveda-Villal{\'o}n, Ikonen, Porras, and G{\'o}mez-P{\'e}rez}{Parkkila et~al\mbox{.}}{2017}]%
        {Parkkila2017AnOF}
\bibfield{author}{\bibinfo{person}{Janne Parkkila}, \bibinfo{person}{Filip Radulovic}, \bibinfo{person}{Daniel Garijo}, \bibinfo{person}{Mar{\'i}a Poveda-Villal{\'o}n}, \bibinfo{person}{Jouni Ikonen}, \bibinfo{person}{Jari Porras}, {and} \bibinfo{person}{Asunci{\'o}n G{\'o}mez-P{\'e}rez}.} \bibinfo{year}{2017}\natexlab{}.
\newblock \showarticletitle{An ontology for videogame interoperability}.
\newblock \bibinfo{journal}{\emph{Multimedia Tools and Applications}}  \bibinfo{volume}{76} (\bibinfo{year}{2017}), \bibinfo{pages}{4981--5000}.
\newblock


\bibitem[\protect\citeauthoryear{Potoniec, Wiśniewski, Ławrynowicz, and Keet}{Potoniec et~al\mbox{.}}{2020}]%
        {POTONIEC2020105098}
\bibfield{author}{\bibinfo{person}{Jedrzej Potoniec}, \bibinfo{person}{Dawid Wiśniewski}, \bibinfo{person}{Agnieszka Ławrynowicz}, {and} \bibinfo{person}{C.~Maria Keet}.} \bibinfo{year}{2020}\natexlab{}.
\newblock \showarticletitle{Dataset of ontology competency questions to SPARQL-OWL queries translations}.
\newblock \bibinfo{journal}{\emph{Data in Brief}}  \bibinfo{volume}{29} (\bibinfo{year}{2020}), \bibinfo{pages}{105098}.
\newblock


\bibitem[\protect\citeauthoryear{Poveda-Villalón, Fernández-Izquierdo, Fernández-López, and García-Castro}{Poveda-Villalón et~al\mbox{.}}{2022}]%
        {POVEDAVILLALON2022LOT}
\bibfield{author}{\bibinfo{person}{María Poveda-Villalón}, \bibinfo{person}{Alba Fernández-Izquierdo}, \bibinfo{person}{Mariano Fernández-López}, {and} \bibinfo{person}{Raúl García-Castro}.} \bibinfo{year}{2022}\natexlab{}.
\newblock \showarticletitle{LOT: An industrial oriented ontology engineering framework}.
\newblock \bibinfo{journal}{\emph{Engineering Applications of Artificial Intelligence}}  \bibinfo{volume}{111} (\bibinfo{year}{2022}), \bibinfo{pages}{104755}.
\newblock


\bibitem[\protect\citeauthoryear{Presutti, Daga, Gangemi, and Blomqvist}{Presutti et~al\mbox{.}}{2009}]%
        {Presutti2009eXtremeDW}
\bibfield{author}{\bibinfo{person}{Valentina Presutti}, \bibinfo{person}{Enrico Daga}, \bibinfo{person}{Aldo Gangemi}, {and} \bibinfo{person}{Eva Blomqvist}.} \bibinfo{year}{2009}\natexlab{}.
\newblock \showarticletitle{EXtreme Design with Content Ontology Design Patterns}. In \bibinfo{booktitle}{\emph{Proceedings of the 2009 International Conference on Ontology Patterns}} \emph{(\bibinfo{series}{WOP'09})}. \bibinfo{publisher}{CEUR-WS.org}, \bibinfo{pages}{83–97}.
\newblock


\bibitem[\protect\citeauthoryear{Reimers and Gurevych}{Reimers and Gurevych}{2019}]%
        {reimers-gurevych-2019-sentence}
\bibfield{author}{\bibinfo{person}{Nils Reimers} {and} \bibinfo{person}{Iryna Gurevych}.} \bibinfo{year}{2019}\natexlab{}.
\newblock \showarticletitle{Sentence-{BERT}: Sentence Embeddings using {S}iamese {BERT}-Networks}. In \bibinfo{booktitle}{\emph{Proceedings of the 2019 Conference on Empirical Methods in Natural Language Processing and the 9th International Joint Conference on Natural Language Processing (EMNLP-IJCNLP)}}. \bibinfo{publisher}{Association for Computational Linguistics}, \bibinfo{pages}{3982--3992}.
\newblock


\bibitem[\protect\citeauthoryear{Ren, Parvizi, Mellish, Pan, van Deemter, and Stevens}{Ren et~al\mbox{.}}{2014}]%
        {Ren2014}
\bibfield{author}{\bibinfo{person}{Yuan Ren}, \bibinfo{person}{Artemis Parvizi}, \bibinfo{person}{Chris Mellish}, \bibinfo{person}{Jeff~Z. Pan}, \bibinfo{person}{Kees van Deemter}, {and} \bibinfo{person}{Robert Stevens}.} \bibinfo{year}{2014}\natexlab{}.
\newblock \showarticletitle{Towards Competency Question-Driven Ontology Authoring}. In \bibinfo{booktitle}{\emph{The Semantic Web: Trends and Challenges}}. \bibinfo{pages}{752--767}.
\newblock


\bibitem[\protect\citeauthoryear{Sequeda, Briggs, Miranker, and Heideman}{Sequeda et~al\mbox{.}}{2019}]%
        {juan2019}
\bibfield{author}{\bibinfo{person}{Juan~F. Sequeda}, \bibinfo{person}{Willard~J. Briggs}, \bibinfo{person}{Daniel~P. Miranker}, {and} \bibinfo{person}{Wayne~P. Heideman}.} \bibinfo{year}{2019}\natexlab{}.
\newblock \showarticletitle{A Pay-as-you-go Methodology to Design and Build Enterprise Knowledge Graphs from Relational Databases}. In \bibinfo{booktitle}{\emph{The Semantic Web -- ISWC 2019}}. \bibinfo{pages}{526--545}.
\newblock


\bibitem[\protect\citeauthoryear{Staab, Studer, Schnurr, and Sure}{Staab et~al\mbox{.}}{2001}]%
        {Staab2001}
\bibfield{author}{\bibinfo{person}{Steffen Staab}, \bibinfo{person}{Rudi Studer}, \bibinfo{person}{Hans-Peter Schnurr}, {and} \bibinfo{person}{York Sure}.} \bibinfo{year}{2001}\natexlab{}.
\newblock \showarticletitle{Knowledge Processes and Ontologies}.
\newblock \bibinfo{journal}{\emph{IEEE Intelligent Systems}} \bibinfo{volume}{16}, \bibinfo{number}{1} (\bibinfo{year}{2001}), \bibinfo{pages}{26–34}.
\newblock


\bibitem[\protect\citeauthoryear{Su{\'a}rez-Figueroa, G{\'o}mez-P{\'e}rez, and Fern{\'a}ndez-L{\'o}pez}{Su{\'a}rez-Figueroa et~al\mbox{.}}{2015}]%
        {SurezFigueroa2015NEON}
\bibfield{author}{\bibinfo{person}{Mari~Carmen Su{\'a}rez-Figueroa}, \bibinfo{person}{Asunci{\'o}n G{\'o}mez-P{\'e}rez}, {and} \bibinfo{person}{Mariano Fern{\'a}ndez-L{\'o}pez}.} \bibinfo{year}{2015}\natexlab{}.
\newblock \showarticletitle{The NeOn Methodology framework: A scenario-based methodology for ontology development}.
\newblock \bibinfo{journal}{\emph{Applied ontology}} \bibinfo{volume}{10}, \bibinfo{number}{2} (\bibinfo{year}{2015}), \bibinfo{pages}{107--145}.
\newblock


\bibitem[\protect\citeauthoryear{Wi{\'s}niewski, Potoniec, {\L}awrynowicz, and Keet}{Wi{\'s}niewski et~al\mbox{.}}{2019}]%
        {wisniewski2019analysis}
\bibfield{author}{\bibinfo{person}{Dawid Wi{\'s}niewski}, \bibinfo{person}{Jedrzej Potoniec}, \bibinfo{person}{Agnieszka {\L}awrynowicz}, {and} \bibinfo{person}{C~Maria Keet}.} \bibinfo{year}{2019}\natexlab{}.
\newblock \showarticletitle{Analysis of Ontology Competency Questions and their formalizations in SPARQL-OWL}.
\newblock \bibinfo{journal}{\emph{Journal of Web Semantics}}  \bibinfo{volume}{59} (\bibinfo{year}{2019}), \bibinfo{pages}{100534}.
\newblock


\end{thebibliography}

\appendix

\end{document}